\documentclass[conference]{IEEEtran}
\IEEEoverridecommandlockouts

\usepackage{cite}
\usepackage{amsmath,amssymb,amsfonts}
\usepackage{graphicx}
\usepackage{textcomp}
\usepackage{xcolor}

\usepackage{algorithm,algpseudocode}
\usepackage{cleveref}
\usepackage{subfigure}
\usepackage{bbm}
\usepackage{mathtools}
\usepackage{comment}
\usepackage{multirow}
\usepackage{url}

\def\BibTeX{{\rm B\kern-.05em{\sc i\kern-.025em b}\kern-.08em
    T\kern-.1667em\lower.7ex\hbox{E}\kern-.125emX}}
\begin{document}

\title{Few-Shot Image-to-Semantics Translation for Policy Transfer in Reinforcement Learning
\thanks{This research is partially supported by the JSPS KAKENHI Grant Number 19H04179, and based on a project, JPNP18002, commissioned by NEDO.
}
}


\author{
\IEEEauthorblockN{Rei Sato\IEEEauthorrefmark{1}\IEEEauthorrefmark{3},
{Kazuto Fukuchi}\IEEEauthorrefmark{2}\IEEEauthorrefmark{3},
{Jun Sakuma}\IEEEauthorrefmark{2}\IEEEauthorrefmark{3} and
{Youhei Akimoto}\IEEEauthorrefmark{2}\IEEEauthorrefmark{3}}
\IEEEauthorblockA{\IEEEauthorrefmark{1}
\textit{Graduate School of Science and Technology}, \textit{University of Tsukuba}, Tsukuba, Japan \\
reisato@bbo.cs.tsukuba.ac.jp\\
\IEEEauthorrefmark{2}
\textit{Faculty of Engineering, Information and Systems}, \textit{University of Tsukuba}, Tsukuba, Japan \\
fukuchi@cs.tsukuba.ac.jp, jun@cs.tsukuba.ac.jp, akimoto@cs.tsukuba.ac.jp\\
\IEEEauthorrefmark{3}
\textit{RIKEN Center for Advanced Intelligence Project}}}

\maketitle

\begin{abstract}
We investigate policy transfer using image-to-semantics translation to mitigate learning difficulties in vision-based robotics control agents. This problem assumes two environments: a simulator environment with semantics, that is, low-dimensional and essential information, as the state space, and a real-world environment with images as the state space. By learning mapping from images to semantics, we can transfer a policy, pre-trained in the simulator, to the real world, thereby eliminating real-world on-policy agent interactions to learn, which are costly and risky. In addition, using image-to-semantics mapping is advantageous in terms of the computational efficiency to train the policy and the interpretability of the obtained policy over other types of sim-to-real transfer strategies. To tackle the main difficulty in learning image-to-semantics mapping, namely the human annotation cost for producing a training dataset, we propose two techniques: pair augmentation with the transition function in the simulator environment and active learning. We observed a reduction in the annotation cost without a decline in the performance of the transfer, and the proposed approach outperformed the existing approach without annotation.
\end{abstract}


\section{Introduction}
Deep reinforcement learning (DRL) has been actively studied for robot control applications in real-world environments because of its ability to train vision-based agents; that is, the robot control actions are output directly from the observed images \cite{corl-kalas, drl1, drl3, large}.
One of the major advantages of vision-based agents in robotics is that camera-captured images can be incorporated into the decision-making of the agent without using a handcrafted feature extractor.

However, allowing vision-based robot control agents to learn by reinforcement learning in the real-world is challenging in terms of risk and cost because it requires a large amount of real-world interactions with unstable robots.
Reinforcement learning involves a learning policy interacting with the environment, and it is theoretically and empirically known that the length of the interaction required for training increases with the dimension of the state space \cite{isa,tassa}.

To address the difficulty associated with reinforcement learning in a real-world environment, methods have been proposed that pre-train a policy on a simulator environment and transfer it to the real-world environment \cite{dr,rcan,darla,aac,mlvr,dawspc,grasp-gan,rl-cycle,retinagan,mdqn,adda-sim2real}.
In this methodology, policies are learned in a simulator, that is, a reinforcement learning environment on a computer that mimics the real-world environment. The policy pre-trained in the simulator is expected to be the optimal policy in the real-world environment.

However, developing a simulator that imitates the real-world environment is not always an easy task. Particularly, because the real world provides image observations, a simulator environment requires a renderer to generate images as states. However, producing a renderer that can generate photorealistic images is fraught with financial and technical difficulties.

In the case that a photorealistic renderer cannot be produced, another style of observations must be adopted as states during the pre-training of the policy in a simulator environment.
Most existing approaches substitute photorealistic observations for non-photorealistic ones using transfer techniques \cite{dr,rcan,darla,aac,mlvr,dawspc,grasp-gan,rl-cycle,retinagan,mdqn,adda-sim2real}.

We investigated a type of transfer strategy called \emph{image-to-semantics} to deal with the absence of a photorealistic renderer, which was created by \cite{zhang}.
In this approach, the \emph{semantics}---low-dimensional and essential information of a state that represents an image---are employed as a form of state observation instead of images in the simulator environment.
The transfer algorithm consists of two steps: pre-training a policy on the simulator environment with semantics as its observation, obtaining a mapping from photorealistic images to their corresponding semantics, and
using the image-to-semantics mapping as a pre-processing component of the policy in the real-world environment. A semantics-based pre-trained policy can be operated in the real-world environment using image observations.
In addition to being a solution to the case without a photorealistic renderer, image-to-semantics mapping has advantages in terms of the computational cost for policy pre-training in the simulator and the interpretability of the acquired policy.

The crucial part of this approach is obtaining the image-to-semantics translation mapping.
To the best of our knowledge, \cite{crar,zhang} are the only studies that have dealt with learning image-to-semantics translation.
We highlight the remaining problems of \cite{crar,zhang}: (1) \cite{crar} used a paired dataset, that is, multiple pairs of images and corresponding semantics, to train the mapping. Considerable human effort is required to make a paired dataset because human annotators provide semantics that represent images. (2) Although the style translation method without a paired dataset \cite{zhang} aims at saving annotation cost, its performance is not often satisfactory owing to the low approximation quality of the image-to-semantics translation mapping, as confirmed in our experiments.

In this study, we tackled learning image-to-semantics translation using a paired dataset; however, we reduced the cost of creating a paired dataset using two strategies: \emph{pair augmentation} and \emph{active learning}.
In our experiments, we confirmed the following claims: first, compared to \cite{crar}, we reduced the cost of making a paired dataset while preserving the performance of the policy transfer. Second, we achieved significantly higher performance than \cite{zhang}, in which a paired dataset was not used, by using a small paired dataset.
For practicality, we conducted experiments under the condition that only inaccurate paired data can be obtained due to various errors, such as annotation errors, and confirmed that the proposed method has a certain robustness against errors.

Our code is publicly available at \url{https://github.com/madoibito80/im2sem}.

\section{Problem Formulation}

\subsection{Markov Decision Process (MDP)}\label{sec:mdp}
We defined a vision-based robotics task in the real world; that is, the real-world environment is a \emph{target MDP}:  $\mathcal{M}^{\tau} = (\mathcal{S}^\tau, \mathcal{A}, p^{\tau}, r^{\tau}, \gamma)$, where $\mathcal{S}^\tau$ is a state space, $\mathcal{A}$ is an action space, $p^\tau: \mathcal{S}^\tau \times \mathcal{A} \times \mathcal{S}^\tau \to \mathbb{R}$ is a transition probability density, $r^\tau:\mathcal{S}^\tau \times \mathcal{A} \times \mathcal{S}^\tau\to \mathbb{R}$ is a reward function, and $\gamma \in [0,1]$ is a discount factor.
Because we assumed that the target MDP is a vision-based task, $\mathcal{S}^{\tau}$ consists of images, and each $s \in \mathcal{S}^\tau$ contains single or multiple image frames.
In standard model-free reinforcement learning (RL) settings, agents can interact with the environment: they observe $s_{t+1} \sim p^\tau(\cdot\mid a_t,s_t)$ and reward $r_t = r^\tau(s_{t+1}, a_t, s_{t})$ by performing action $a_t$ at state $s_t$, which is internally preserved in the environment at timestep $t$; after the transition, $s_{t+1}$ is stored in the environment.
However, there are concerns in terms of the risk and cost associated with learning a policy through extensive interaction with $\mathcal{M}^\tau$.

To reduce the risk and cost of training a policy in the target MDP, we pre-trained a policy on a simulator environment, called the \emph{source MDP}: $\mathcal{M}^{\sigma} = (\mathcal{S}^\sigma, \mathcal{A}, p^{\sigma}, r^{\sigma}, \gamma)$.
Note that the action space $\mathcal{A}$ is the same between the two MDPs. 
In contrast, the state space $\mathcal{S}^\sigma$, the transition probability density $p^\sigma: \mathcal{S}^\sigma\times\mathcal{A}\times\mathcal{S}^\sigma\to\mathbb{R}$, and the reward function $r^\sigma:\mathcal{S}^\sigma\times\mathcal{A}\times\mathcal{S}^\sigma\to\mathbb{R}$ are different from those of the target MDP.
We assumed that because we considered robotics tasks, the deterministic transition function $Tr^\sigma(s,a) = s' \sim p^\sigma(\cdot \mid a,s)$ could be defined in the simulator environment and $p^\sigma$ resembled a Dirac delta distribution.

The source state space $\mathcal{S}^\sigma$ corresponded to a \emph{semantic space}, that is, each $s \in \mathcal{S}^\sigma$ was semantic information.
For example, consider a robot-arm grasp task; each $s \in \mathcal{S}^\tau$ is a single or multiple image frame showing a robot arm and objects to be grasped.
Each $s \in \mathcal{S}^\sigma$ consists of semantics such as $xyz$-coordinates of the end-effector and target objects and angles of joints.

The source MDP and target MDP are expected to have some structural correspondence.
Here, we describe our assumptions regarding the relations of the two MDPs.
We assumed the existence of a function $F:\mathcal{S}^{\tau}\to\mathcal{S}^{\sigma}$ satisfying the following conditions:

\emph{Transition Condition: }
For all $(s', a, s) \in \mathcal{S}^{\tau} \times \mathcal{A} \times \mathcal{S}^{\tau}$, $p^{\sigma}(F(s') \mid a, F(s)) = \int_{\bar{s} \in \bar{\mathcal{S}}} p^{\tau}(\bar{s} \mid a, s) \mathrm{d}\bar{s}$, where $\bar{\mathcal{S}} = \{\bar{s}\in\mathcal{S}^\tau \mid F(\bar{s}) = F(s')\}$.

\emph{Reward Condition: }
For all $(s', a, s) \in \mathcal{S}^{\tau} \times \mathcal{A} \times \mathcal{S}^{\tau}$, $r^{\sigma}(F(s'), a, F(s)) = r^{\tau}(s',a,s)$.

In the above conditions, $F$ is considered an oracle that takes an image and outputs corresponding semantics; that is, $F$ is the true image-to-semantics translation mapping.
In the transition condition, $\bar{\mathcal{S}}$ is a set of images that has common semantics $F(s')$.
Imagine the transition from $s\in\mathcal{S}^\tau$ to $s'\in\mathcal{S}^\tau$ with action $a \in \mathcal{A}$ in the target MDP, the transition condition holds $F(s') = Tr^\sigma(F(s),a)$.
The reward condition indicates that a reward for this transition $r^\tau(s',a,s)$ equals the one for a transition from $F(s)\in\mathcal{S}^\sigma$ to $F(s')\in\mathcal{S}^\sigma$ with the action $a$ in the source MDP.

\subsection{Transfer via Image-to-Semantics}

\begin{figure}[t]
\centerline{\includegraphics[bb=0 0 446 211, width=0.48\textwidth]{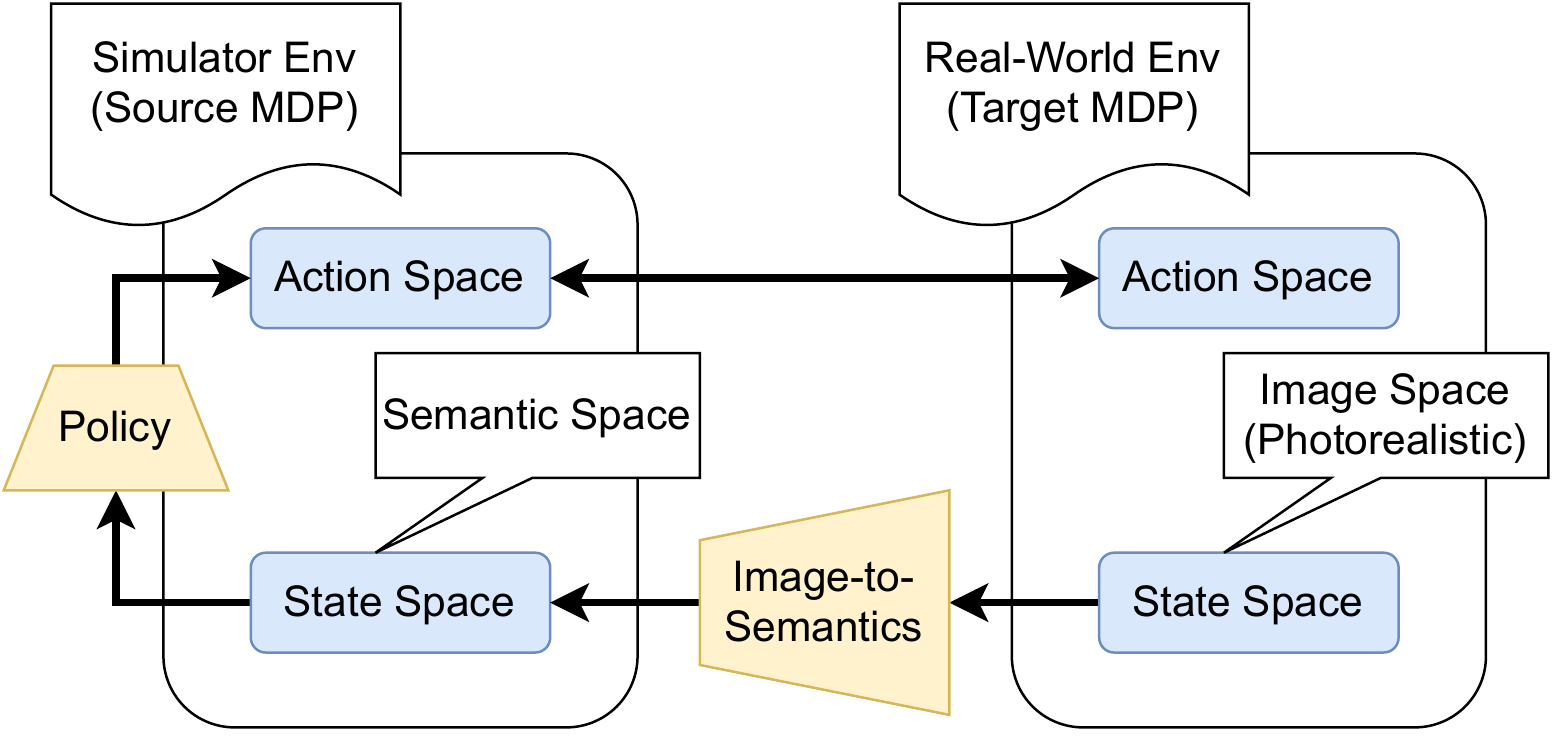}}
\caption{Illustration of transfer via image-to-semantics. We approximated the image-to-semantics translation mapping $F$ as $\hat{F}$. Because the action space was common to both MDPs, we operated the composite of the source policy $\pi^\sigma$ and approximated image-to-semantics translation mapping $\hat{F}$, that is, $\pi^\sigma\circ \hat{F}$ in the target MDP.}
\label{fig:illu3}
\end{figure}

\subsubsection{Policy Transfer}
The objective of RL is the expectation of the discounted cumulative reward: 
\begin{equation}\label{eq:rl-obj}\textstyle
J(\pi; p, r, \gamma, p_{0}) = \mathbb{E}_{\pi, p, p_{0}}\left[\sum_{t=0}^{\infty} \gamma^{t} r(s_{t+1}, a_t, s_{t})\right]
\end{equation}
and maximizing it w.r.t.~$\pi$.
Here, $\pi: \mathcal{S} \times \mathcal{A} \to \mathbb{R}$ is a policy, that is, a conditional distribution of $a_{t}$ given $s_t$, and $p_{0}$ is the distribution of the initial state $s_{0}$ over the state space.
Our objective was to obtain a well-trained policy on the target MDP: $\pi^\tau = \mathrm{arg~max}_{\bar{\pi}^\tau} J(\bar{\pi}^{\tau}; p^\tau, r^\tau, \gamma, p^{\tau}_{0})$.

Under the situation in which the transition and reward conditions mentioned above hold for some $F$, we can replace $\pi^\tau$ by $\pi^\sigma \circ F$, where $\pi^\sigma$ is a well-trained policy on the source MDP, that is, $\pi^\sigma = \mathrm{arg~max}_{\bar{\pi}^\sigma} J(\bar{\pi}^{\sigma}; p^\sigma, r^\sigma, \gamma, p^{\sigma}_{0})$. Solving this maximization by RL requires sole interaction with $\mathcal{M}^\sigma$ instead of $\mathcal{M}^\tau$.
As noted, interactions with $\mathcal{M}^\tau$ require real-world operations; however, interactions with $\mathcal{M}^\sigma$ are performed on the simulator, which is cost-effective.

Based on this property, we studied the following transfer procedure: pre-train $\pi^\sigma$ on $\mathcal{M}^\sigma$, approximate $F$ as $\hat{F}$, and output the target agent $\pi^\sigma\circ \hat{F}$. This procedure was investigated by \cite{zhang}.
\Cref{fig:illu3} illustrates the transfer via image-to-semantics.

\subsubsection{Advantages}\label{sec:adv}
The above-mentioned transfer strategy, that is, transfer via image-to-semantics, has the following three advantages over approaches using a renderer in the source MDP shown in \Cref{tab:related}.
First, a renderer is not required.
Existing methods that use a renderer generally aim to transfer an agent based on non-photorealistic images in a simulator to photorealistic images in the real world \cite{dr,rcan,darla,aac,mlvr,dawspc,grasp-gan,rl-cycle,retinagan,mdqn,adda-sim2real}.
Therefore, they require the preparation of a renderer on the simulator to generate non-photorealistic images as state observations.
Transfer via image-to-semantics performs similar transfer learning; however, it does not require a renderer because the source MDP has a semantic space as its state space. This can reduce the development cost of the simulator for some tasks.
Second, because semantics are low-dimensional variables compared to images, we can improve the sample efficiency required to train the policy $\pi^\sigma$ on $\mathcal{M}^\sigma$ \cite{isa,tassa}.
Learning vision-based agents are generally associated with large computational costs, even on a simulator \cite{icml-curl}, but
transfer via image-to-semantics is relatively lightweight in this respect and occasionally allows a human to design the policy.
Third, using semantics as an intermediate representation of the target agent contributes to its high interpretability because of the low-dimensionality and interpretability of semantics.
Similar to \cite{crar,acml-yang}, because the real-world agent $\pi^\sigma \circ \hat{F}$ can be separated into two components, which are independently trained, it is easier to assess than one trained in an end-to-end manner.

\subsection{Resource Strategy}\label{sec:resource}
In this section, in addition to the two MDP environments, we define resources that can be used to approximate $F$.

\subsubsection{Transition Function}
In the target MDP, the state transition result $s_{t+1}$ due to the selected action $a_t$ can be observed only for state $s_t$ stored inside the environment.
In contrast, in the source MDP, we assumed that the state transition result for any $s \in \mathcal{S}^\sigma$ could be observed, replacing the $s_t$ stored inside the environment with $s$.
This is because the actual state transition probability $p^\tau$ in the target MDP is a physical phenomenon in the real world, but the state transition rule $Tr^\sigma$ in the source MDP is a black-box function on the computer.

\subsubsection{Offline Dataset}
The offline dataset comprised observations of the target MDP, that is, $\mathcal{T}^\tau = \{(s_t, a_t, \mathbbm{1}_{\mathrm{end}}(s_{t+1})) \in \mathcal{S}^\tau \times \mathcal{A} \times \{0,1\}\}_{t}$, where $\mathbbm{1}_{\mathrm{end}}(s_{t+1}) = 1$ represents that $s_{t+1}$ corresponding to a terminal state; otherwise, $0$.
Note that successive indices in the offline dataset shared the same context of the episode, except at the end of the episode.
$\mathcal{T}^\tau$ can be obtained before training starts and is collected by a behavior policy.
Because the offline dataset can be reused for any trial and be obtained by a safety-guaranteed behavior policy, we assumed it could be created at a relatively low cost.

We solely used the offline dataset for supervised and unsupervised learning purposes.
If offline reinforcement learning is executed, the vision-based agent can be trained directly without approximating $F$. However, training a vision-based agent using an offline dataset by reinforcement learning requires large-scale trajectories in the scope of millions \cite{icml20-off}.
In this study, we considered situations in which the total number of timesteps in the offline dataset was limited, for example, less than 100k timesteps.

We did not need to generate reward signals while collecting the offline dataset.
World models \cite{worldmodels} have been studied for the procedure: approximate MDP $\mathcal{M}$ as $\hat{\mathcal{M}}$ using an offline dataset of $\mathcal{M}$; train a policy by reinforcement learning by interacting with the approximated environment $\hat{\mathcal{M}}$ instead of interacting with the original environment $\mathcal{M}$.
One could imagine that we could replace interactions with the target MDP by interactions with the approximated one.
However, to accomplish this, we must observe signals regarding reward in the real world while collecting the offline dataset, and we must approximate a reward function that is often sparse; both of these are not always easy \cite{idapt}.
Therefore, we did not consider approximating the target MDP and did not assume the reward was contained in $\mathcal{T}^\tau$.

\subsubsection{Paired Dataset}
The paired dataset $\mathcal{P}$ consisted of multiple pairs of target state observations and their corresponding source state observations.
Let $\mathcal{I}$ denote the set of indices that indicate the position of the offline dataset.
Using the true image-to-semantics translation mapping $F$, we can denote $\mathcal{P} = \{(F(s_i), s_i) \mid (s_i, a_i, e_i) \in \mathcal{T}^\tau, i \in \mathcal{I}\}$.
Under practical situations, querying $F$ equals annotating corresponding semantics to the images of the indices $\mathcal{I}$ in the offline dataset $\mathcal{T}^\tau$ by human annotators.
Because of its annotation cost, we assumed the size of the paired dataset $\lvert\mathcal{I}\rvert$ to be significantly smaller than that of the offline dataset, for example, $\lvert\mathcal{I}\rvert \leq 100$.

\section{Related Work}
\label{sec:related}

\begin{table}[t]
\centering
\caption{Related policy transfer methods for observation style shift.
Each method requires different resources: renderer, offline dataset (OFF), and paired dataset (PAIR).}\label{tab:related}
\begin{tabular}{l|c|c|c}
      \hline
    Method & Renderer & OFF & PAIR \\
        \hline
    Tobin et al.\cite{dr} & \checkmark & &  \\
    RCAN\cite{rcan} & \checkmark & &  \\
    DARLA\cite{darla} & \checkmark & & \\
    Pinto et al.\cite{aac} & \checkmark & & \\
    MLVR\cite{mlvr} & \checkmark & & \\
    Tzeng et al.\cite{dawspc} & \checkmark & \checkmark & \\
    GraspGAN\cite{grasp-gan} & \checkmark & \checkmark \\
    RL-CycleGAN\cite{rl-cycle} & \checkmark & \checkmark \\
    RetinaGAN\cite{retinagan} & \checkmark & \checkmark & \\
    MDQN\cite{mdqn} & \checkmark & \checkmark & \checkmark \\
    ADT\cite{adda-sim2real} & \checkmark & \checkmark & \checkmark \\
    \hline
    Zhang et al.\cite{zhang} & & \checkmark & \\
    CRAR\cite{crar} & & \checkmark & \checkmark \\
    \textbf{Ours} & & \checkmark & \checkmark \\
      \hline
\end{tabular}
\end{table}

We introduced some existing sim-to-real transfer methods that use a non-photorealistic renderer on the simulator.
\Cref{tab:related} lists the transfer methods that do not require on-policy interaction in the target MDP, assuming vision-based agents.
The main difficulty tackled by these methods was the absence of a photorealistic renderer on the simulator. In the real world, images captured by a camera are input to the agent; however, generating photorealistic images on the simulator is generally difficult because it requires developing a high-quality renderer.

In \cite{dr,rcan,darla,aac,mlvr,driving}, the algorithms learned policies or intermediate representations that were robust to changes in image style using a non-photorealistic renderer.
Thus, these algorithms were expected to perform well even when a photorealistic style was applied in a real-world environment.
In particular, the domain randomization technique has been widely used \cite{dr,rcan,darla,aac}.

\subsection{Transfer via Image-to-Image Translation}
In contrast to the above methods, \cite{dawspc, grasp-gan, rl-cycle, retinagan, adda-sim2real, mdqn, zhang, crar} aimed to perform style translation mapping among specific styles.
To accomplish this, these methods required an offline dataset of the target MDP.
Because these methods followed the principle of collection without execution of on-policy interaction, the offline dataset could be collected by a safety-guaranteed policy.
Unsupervised style translation, such as domain adaptation \cite{adda} and CycleGAN \cite{CycleGAN2017}, are often used to change the styles for state-of-the-art methods \cite{grasp-gan, rl-cycle, retinagan, adda-sim2real, zhang, gamrian, idapt}. Using this translation mapping as a pre-processing function of the target agent, the pre-trained policy can determine actions in the same image style as the source MDP in the target MDP.

However, domain adaptation and cycle-consistency \cite{CycleGAN2017} only have a weak alignment ability \cite{zhang}, and some existing methods use paired datasets to properly transfer styles \cite{crar, adda-sim2real, mdqn}.
Therefore, these two datasets have been widely employed in previous studies and can be assumed to be a common setting.

The similarity of transfer via image-to-semantics and image-to-image is that they train style translation mapping $\hat{F}$ among the source and target state spaces that preserves essential information; furthermore, the agent is the composite $\pi \circ \hat{F}$, where $\pi$ is a policy.

Again, the above methods use a non-photorealistic renderer on the simulator.
Thus, these methods cannot be compared with transfer via image-to-semantics, as explained in \Cref{sec:adv}.

\subsection{Learning Image-to-Semantics}
Previous studies have used semantics in the source MDP \cite{zhang, aac, adda-sim2real, mdqn, dawspc}. 
An important perspective on the applicability of these methods to image-to-semantics is whether they use a renderer on the simulator, as shown in \Cref{tab:related} and as discussed in \Cref{sec:adv}.
Because methods using a renderer assume that the source state space is an image space, image-to-semantics is beyond their scope, and it is not certain that their mechanism will be successful in image-to-semantics. For example, CycleGAN, which has been successfully used for image-to-image learning, failed in image-to-semantics \cite{zhang}. 
In this regard, we refer to \cite{zhang}, an unpaired method that applies the findings from image-to-image to image-to-semantics.
In addition, \cite{crar} is compared as a representative method that uses a paired dataset as in this study.

\subsubsection{CRAR}
We refer to Section 4.4 of CRAR \cite{crar} as a baseline of image-to-semantics learning.
They described the following policy transfer strategy: pre-train a source state encoder $E^\sigma:\mathcal{S}^\sigma \to \mathcal{Z}$, where $\mathcal{Z}$ is a latent space of the encoder; train the source policy $\pi^\sigma: \mathcal{Z}\to\mathcal{A}$; and train a target state encoder $E^\tau: \mathcal{S}^\tau\to\mathcal{Z}$ with regularization term $\sum_{(s^\sigma, s^\tau) \in \mathcal{P}} \lVert E^\sigma(s^\sigma) - E^\tau(s^\tau)\rVert_2^2$, where $\mathcal{P}$ is a paired dataset. Then, the target agent is the composite $\pi^\sigma \circ E^\tau: \mathcal{S}^\tau\to\mathcal{A}$.
Here, $E^\tau$ can be regarded as a style translation mapping.
Note that they only performed this experiment in the setting where $\mathcal{S}^\sigma$ and $\mathcal{S}^\tau$ are both image spaces; however, it can be applied easily where $\mathcal{S}^\sigma$ is the semantic space.

\subsubsection{Zhang et al.}
We referred to the \emph{cross-modality} setting of their experiment as our baseline for image-to-semantics \cite{zhang}.
This setting is the same as the transfer via image-to-semantics.

There remain some challenges in \cite{crar, zhang}.
For \cite{zhang}, the human annotation cost was eliminated because they did not use a paired dataset.
However, the loss function defined by \cite{zhang} for unpaired image-to-semantics style translation will not necessarily provide a well-approximated $F$.
Therefore, we decided to use a paired dataset to efficiently supervise the loss function as performed in \cite{crar}, but with a paired dataset smaller than \cite{crar}.

\section{Methodology}

\begin{figure*}[t]
\centerline{\includegraphics[bb=0 0 730 228, width=0.98\textwidth]{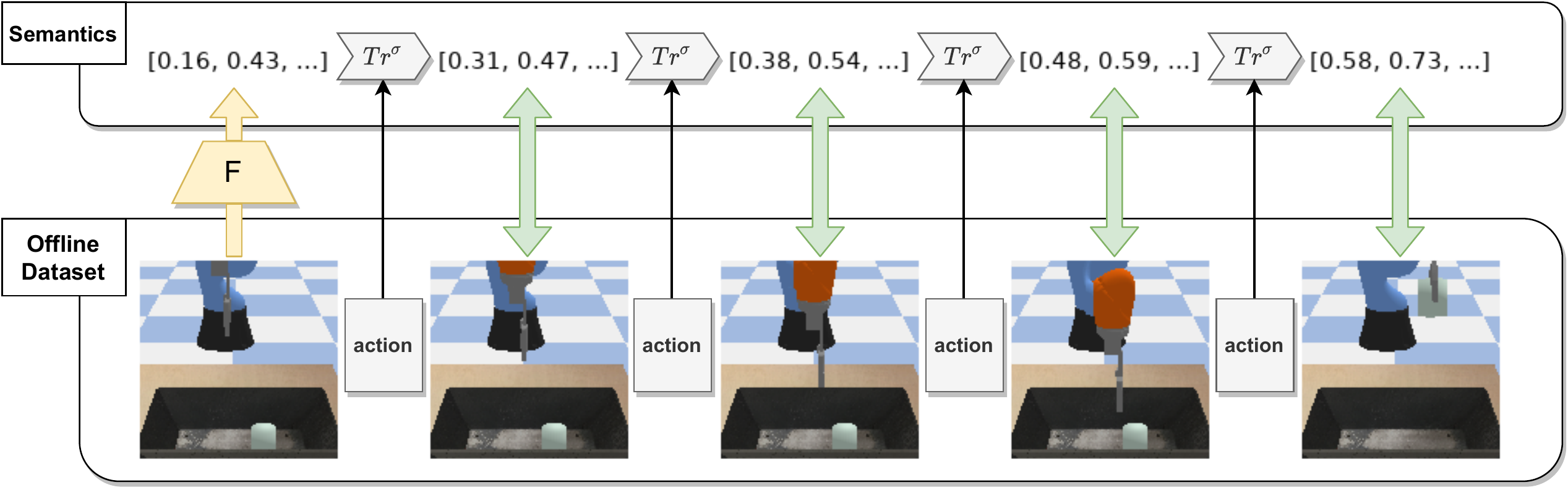}}
\caption{Illustration of pair augmentation.
Oracle $F$ generates semantics corresponding to a particular image in the offline dataset. 
The next state in semantics is computed using the transition function $Tr^\sigma$ with the current semantics along with the action taken while collecting the offline dataset.
This allows us to obtain semantics corresponding to the image at the next timestep in the offline dataset without any annotation costs. 
Augmented pairs with green dual directional arrows were stored in $\mathcal{P}'$.
In this figure, note that rendered (non-photorealistic) images are shown in the offline dataset, but in reality, camera-captured (photorealistic) images are contained.}
\label{fig:illu}
\end{figure*}

Our approach approximates the image-to-semantics translation $F$ using an offline dataset $\mathcal{T}^\tau$. 
Similar to \cite{crar}, we used a paired dataset $\mathcal{P} = \{(F(s_i), s_i) \mid (s_i, a_i, e_i) \in \mathcal{T}^\tau, i \in \mathcal{I}\}$, which was constructed by querying $F(s_i)$ to human annotators for an image observation of target MDP $s_i \in \mathcal{S}^\tau$ included in $\mathcal{T}^\tau$.
We incorporated two main ideas to reduce the annotation cost. \emph{Pair augmentation} generates an augmented paired dataset $\mathcal{P}'$ using an offline dataset $\mathcal{T}^\tau$.
\emph{Active learning} defines $\mathcal{I}$, that is, it selects a subset of $\mathcal{T}^\tau$ to be annotated to construct $\mathcal{P}$ (\Cref{pseudo-algo2}). 
We present an overall procedure of our method in \Cref{algo:overall}.

We assumed that we have an offline dataset $\mathcal{T}^\tau$, which comprises multiple episodes in the target MDP. Let $\mathcal{O}$ denote the set of indices corresponding to the beginning of an episode in $\mathcal{T}^\tau$, that is, $\mathcal{O} = \{0\}\cup\{i \mid 0 < i < \lvert \mathcal{T}^\tau \rvert \text{ and } e_{i-1} = 1 \text{ for } (s^\tau_{i-1}, a_{i-1}, e_{i-1}) \in \mathcal{T}^\tau\}$, where $e_i$ is the indicator: when timestep $i$ is the end of an episode then $e_i = 1$.
For each $i \in \mathcal{O}$, let $\mathcal{E}_{i} = \{t \mid 1\leq t \leq \min(\{k \mid k \geq 1,\ e_{i+k} = 1\})\}$. Then, for each $i \in \mathcal{O}$, a subsequence of $\mathcal{T}^\tau$ starting from timestep $i$ and ending at $i + \lvert \mathcal{E}_i \rvert$ corresponds to an episode.

\begin{algorithm}[t]
  \caption{Overall Procedure}
\label{algo:overall}
  \begin{algorithmic}[1]
    \Require Source MDP $\mathcal{M}^\sigma$, Offline dataset $\mathcal{T}^{\tau}$, Oracle $F$

    \State Train source MDP's policy $\pi^\sigma$ on $\mathcal{M}^\sigma$
    \State Train VAE encoder $E^\tau$ using $\mathcal{T}^{\tau}$
    \State Determine indices $\mathcal{I}$ by active learning (\Cref{pseudo-algo2}) using $E^{\tau}$, $\mathcal{T}^\tau$
    \State Create $\mathcal{P}$ for $\mathcal{I}$, $\mathcal{T}^\tau$ by oracle (human annotator) $F$
    \State Create augmented pairs $\mathcal{P}'$ using $\mathcal{P}$, $\mathcal{T}^\tau, Tr^\sigma\in\mathcal{M}^\sigma$
    \State Train $\hat{F}$ by minimizing \Cref{eq:crar}
    \Ensure Target MDP's agent $\pi^\sigma \circ \hat{F}$
  \end{algorithmic}
\end{algorithm}

\subsection{Pair Augmentation by Transition Function}
The objective of \emph{pair augmentation} is to construct artificial paired data $\mathcal{P}'$ such that $s^\sigma \approx F(s^\tau)$ for $(s^\sigma , s^\tau) \in \mathcal{P}'$ and $s^\tau \in \mathcal{T}^\tau$.
Using an augmented paired dataset, we aimed to obtain $\hat{F}$ that approximates $F$ by minimizing the loss
\begin{equation}
\mathcal{L}(\hat{F}, \mathcal{P}\cup\mathcal{P}') = \frac{1}{\lvert\mathcal{P}\cup\mathcal{P}'\rvert} \sum_{(s^\sigma, s^\tau)\in\mathcal{P}\cup\mathcal{P}'} \lVert s^\sigma - \hat{F}(s^\tau) \rVert_2^{2} \enspace.\label{eq:crar}
\end{equation}
Note that CRAR \cite{crar} adopts $\mathcal{L}(\hat{F}, \mathcal{P})$ instead of $\mathcal{L}(\hat{F}, \mathcal{P}\cup\mathcal{P}')$.

Our principle is as follows. Let $\mathcal{I} \subseteq \mathcal{O}$ be a subset of indices corresponding to the beginning of the episodes in $\mathcal{T}^\tau$.
Suppose we have a paired dataset $\mathcal{P}$ constructed by querying semantics $s_i^\sigma = F(s_i^\tau)$ corresponding to images $s_i^\tau$ in $\mathcal{T}^\tau$ for time index $i \in \mathcal{I}$.
Although semantics $s_{i+1}^\sigma$ representing an image of the next timestep $s^\tau_{i+1}$ in $\mathcal{T}^\tau$ is unknown, because of the transition condition given in \Cref{sec:mdp} and deterministic transition, it equals $s_{i+1}^\sigma = Tr^\sigma(s_{i}^\sigma, a_{i})$, where $a_i$ is the action taken at timestep $i$ when collecting the offline dataset $\mathcal{T}^\tau$ and is included in $\mathcal{T}^\tau$.
In reality, because human annotations and state transition contain errors as compared to the truth, the generated semantics $s^\sigma_{i+1}$ do not exactly represent the image $s^\tau_{i+1}$. However, even with errors in $F$ and $Tr^\sigma$, it is expected that the generation of the above semantics is a valuable approximation.
By recursively applying the above generation, we obtained the augmented paired dataset $\mathcal{P}'$.

Formally, $\mathcal{P}'$ was constructed as follows: 
For each index $i \in \mathcal{I}$, we defined a sequence $\{\widehat{s}_{i+t}^\sigma\}_{t \in \mathcal{E}_{i}}$ as $\widehat{s}_{i}^\sigma = s_{i}^\sigma$ (contained in $\mathcal{P}$) and $\widehat{s}_{i+t}^\sigma = Tr^\sigma(\widehat{s}_{i+t-1}^\sigma, a_{i+t-1})$ for $t \in \mathcal{E}_{i}$, where $a_{i+t-1}$ are contained in $\mathcal{T}^\tau$.
The augmented paired dataset is then $\mathcal{P}' = \{(\widehat{s}^{\sigma}_{i+t}, s^{\tau}_{i+t})\}_{i\in\mathcal{I},t\in\mathcal{E}_{i}}$, where $s^{\tau}_{i+t}$ is contained in $\mathcal{T}^\tau$. 
Thus, we could construct an augmented paired dataset $\mathcal{P}'$ of size $\lvert \mathcal{P}' \rvert = \sum_{i \in \mathcal{I}} \lvert \mathcal{E}_i\rvert$ from the paired dataset $\mathcal{P}$ of size $\lvert \mathcal{P} \rvert = \lvert \mathcal{I} \rvert$.

\Cref{fig:illu} illustrates the pair augmentation scheme.

The reason why $\mathcal{I}$ was a subset of episode start indices $\mathcal{O}$ rather than $\mathcal{I} \subseteq \{j \mid 0\leq j < \lvert \mathcal{T}^\tau \rvert, j\in \mathbb{Z}\}$ was to maximize the size of augmented pairs $\lvert \mathcal{E}_i \rvert$.
In other words, because we could augment $s^\sigma_i = F(s^\tau_i)$ until the end of the episode including $s^\tau_i$, to maximize $\lvert \mathcal{P}\cup\mathcal{P}' \rvert$, human annotations should be conducted at the beginning of an episode of $\mathcal{T}^\tau$.

\subsection{Active Learning for Pair Augmentation}
To select episodes for annotation, that is, decide $\mathcal{I}$, we incorporated the idea of diversity-based \emph{active learning (AL)} \cite{vaal, al-flp, al-flp2}.
Their motivation was to select dissimilar samples to effectively reduce the approximation error.
Intuitively, if $\mathcal{P}\cup\mathcal{P}'$ has many similar pairs, they might have a similar effect on training $\hat{F}$; this may lead to a waste in annotation cost.
Therefore, we attempted to select episodes (indexed by $\mathcal{I} \subset \mathcal{O}$) to be annotated to ensure the inclusion of diverse pairs.

We successively selected the episode to annotate, and we called each selection step the $n$-th round. 
For $i \in \mathcal{O}$, let $B_i = \{s_{i+t}^{\tau}\}_{t \in \{0\} \cup \mathcal{E}_i}$ be a set of target state observations present in the episode starting at timestep $i \in \mathcal{O}$. We referred to it as \emph{batch}. 
Let $\mathcal{I}_{n-1}$ be the set of selected indices before the $n$-th round, and let $S_{n-1} = \bigcup_{k \in \mathcal{I}_{n-1}} B_k$ be a set of all the state vectors in the episodes selected before the $n$-th round.
Let $d: \mathcal{S}^\tau \times \mathcal{S}^\tau \to \mathbb{R}$ be some appropriate distance measure.
In the $n$-th round, a batch was selected based on the following two diversity measures: 
The \emph{inter batch diversity}
\begin{equation}
f_{\mathrm{inter}}(B_i,S_{n-1}) = \sum_{s^\tau \in B_i} \min_{s_{j}^\tau \in S_{n-1}} d(s^\tau, s_{j}^\tau)
\end{equation}
can evaluate the dissimilarity of $B_i$ and $S_{n-1}$.
The batch with the greatest $f_{\mathrm{inter}}$ was considered to be the most dissimilar batch against the pre-selected batches.
The \emph{intra batch diversity}
\begin{equation}
f_{\mathrm{intra}}(B_i) = \sum_{s^\tau_{p} \in B_i} \sum_{s^\tau_{q} \in B_i} d(s_{p}^\tau, s_{q}^\tau)
\end{equation}
can evaluate the dissimilarity of the states inside $B_i$.
The batch with the greatest $f_{\mathrm{intra}}$ was considered to contain the most diverse states.

\begin{algorithm}[t]
  \caption{Active Learning}
\label{pseudo-algo2}
  \begin{algorithmic}[1]
    \Require Trained VAE encoder $E^\tau$, Offline dataset $\mathcal{T}^{\tau}$
    \State Initialize $\mathcal{I}_0 = \{c\}_{|c \sim \mathrm{Uniform}(\mathcal{O})}$
    \For {$1 \leq n < N$}
    \Comment{$n$-th round}
    \State Set $S_{n-1} = \bigcup_{k \in \mathcal{I}_{n-1}} B_k$
    \State Measure $f_{\mathrm{inter}}(B_i, S_{n-1})$ for all $i\in\mathcal{O}$
    \State Pick top $b\%$ of indices in terms of $f_{\mathrm{inter}}$ as $\mathcal{Q}$
    \State Measure $f_{\mathrm{intra}}(B_i)$ for all $i \in \mathcal{Q}$
    \State Pick the index $c$ from $\mathcal{Q}$ with the greatest $f_{\mathrm{intra}}$
    \State Set $\mathcal{I}_{n} = \{c\} \cup \mathcal{I}_{n-1}$
    \EndFor
    \Ensure Indices $\mathcal{I}_{N-1}$ (with the size of $N$) as $\mathcal{I}$
  \end{algorithmic}
\end{algorithm}

We selected a batch that maximizes the above two diversity measures; we performed a bi-objective optimization for selection.
To avoid overemphasizing one measure over the other, we employed two separate single-objective optimizations for each measure.
In each round, we picked up indices of batches with $f_{\mathrm{inter}}$ in the top $b\%$ ($b=10$ in our experiments) from unselected episodes as $\mathcal{Q}$, and subsequently, selected the batch with the greatest $f_{\mathrm{intra}}$ from $\mathcal{Q}$.
$\mathcal{I}_0$ was initialized with the episode sampled from $\mathcal{O}$ uniformly at random.

\subsection{Representation Learning Using Offline Dataset}
For $d: \mathcal{S}^\tau \times \mathcal{S}^\tau \to \mathbb{R}$ to be a reasonable distance measure in the image space, we employed a VAE encoder \cite{vae}: $E^\tau: \mathcal{S}^\tau \to \mathcal{Z}$. It stochastically outputs a latent vector $z \in \mathcal{Z}$ for $s^\tau \in \mathcal{S}^\tau$.
The distance between two states $s_p^\tau \in \mathcal{S}^\tau$ and $s_q^\tau \in \mathcal{S}^\tau$ was given by the Euclidean distance between the mean vectors for their latent representations, that is, $d(s_p^\tau, s_q^\tau) = \lVert \mathbb{E}[E^\tau(s_p^\tau)] - \mathbb{E}[E^\tau(s_q^\tau)]\rVert_{2}$.
We trained $E^\tau$ using all states in the offline dataset $\mathcal{T}^\tau$ before performing the active learning procedure.

We used the states contained in $\bigcup_{i\in\mathcal{I}}B_i$ in training $\hat{F}$ by \Cref{eq:crar}; however, the remaining $\bigcup_{i\in\mathcal{O}\setminus\mathcal{I}}B_i$ were not used.
In order to use it, we included $E^\tau$ as a feature extractor for $\hat{F}$ by receiving the benefit of representation learning for downstream tasks.
We modeled $\hat{F}=\phi\circ E^\tau$, and we trained $\phi$ by \Cref{eq:crar}, whereas $E^\tau$ was fixed.

\section{Experiments}
We aimed to verify the following two claims: (1) the proposed paired augmentation and AL reduces the annotation cost for approximating $\hat{F}$ while maintaining its performance level; and (2) the paradigm with the paired dataset performs better than the method without paired datasets.

\subsection{Evaluation Metrics}
\subsubsection{Policy Performance (PP)}
The most important evaluation metric for $\hat{F}$ is the expected cumulative reward of the target agent using \Cref{eq:rl-obj}:
\begin{equation}
\mathrm{PP}(\hat{F}; \pi^\sigma,\mathcal{M}^\tau) = J(\pi^{\sigma}\circ\hat{F}; p^\tau, r^\tau, \gamma, p^{\tau}_{0}) \, .
\end{equation}
In our experiments, we approximated it by averaging the cumulative reward of 50 episodes with $\gamma=1$.
This metric was commonly used in \cite{crar,zhang}.

\subsubsection{Matching Distance (MD)}
Because our technical contribution was mainly to approximate $F$, we used the following empirical approximation error:
\begin{equation} \label{eq:md}
\mathrm{MD}(\hat{F}; \mathcal{T}, F) = \frac{1}{\lvert\mathcal{T}\rvert} \sum_{(s^{\tau}, a, e) \in \mathcal{T}} \lVert F(s^\tau) - \hat{F}(s^\tau)\rVert_2^2,
\end{equation}
where $\mathcal{T}$ is a trajectory collected by a behavior policy in the target MDP, which is not used  for learning $\hat{F}$.
Unfortunately, in a real-world environment, evaluating \Cref{eq:md} for a large size of $\mathcal{T}$ is challenging because $F$ requires human annotation.
To enable MD in our experiment, we performed experiments using the simulator for both the source MDP and target MDP. We adopted the rendered image space as the state space of the target MDP.
Because both semantics and images were generated in the simulator, $F$ was freely available to calculate \Cref{eq:md}.
A similar metric to \Cref{eq:md} was used in \cite{zhang}.

\subsection{Environment}
We evaluated the proposed approach on three environments.

\subsubsection{ViZDoom Shooting (Shooting)}
ViZDoom Shooting \cite{vizdoom} is a first-person view shooter task, in which an agent obtains 64$\times$64 RGB images from the first-person perspective in the target MDP.
The agent can change its $x$-coordinate by moving left and right in the room and attacking forward ($\lvert\mathcal{A}\rvert=3$).
An enemy spawns with a random $x$-coordinate on the other side of the room at the start of the episode and does not move or attack.
The agent can destroy the enemy by moving to the front of it and shooting it; time to destruction is directly related to the reward.
Semantics are the $x$-coordinates of the agent and the enemy; hence, $\mathcal{S}^\sigma$ is a 2-dimensional space.
The maximum timesteps is 50 for each episode.
The behavior policy to collect the offline dataset $\mathcal{T}^\tau$ is a random policy, and $\mathcal{T}^\tau$ consists of 200 episodes, that is, 10k timesteps in total.

\subsubsection{PyBullet KUKA Grasp (KUKA)}
This is a grasp task using PyBullet's KUKA iiwa robot arm \cite{pybullet}. Success is achieved by manipulating the end-effector of the robot arm and lifting a randomly placed cylinder. The semantics are the $xyz$-coordinate and the 3-dimensional Euler angle of the end-effector and the $xyz$-coordinate of the cylinder; hence, $\mathcal{S}^\sigma$ is a 9-dimensional space.
We used the rendered 64$\times$64 RGB images captured from three different viewpoints simultaneously as the state observations in the target MDP.
The total timesteps per episode is fixed to 40.
The behavior policy to collect $\mathcal{T}^\tau$ is a random policy, and $\mathcal{T}^\tau$ comprised 250 episodes, that is, 10k timesteps in total.

\subsubsection{PyBullet HalfCheetah-v0 (HalfCheetah)}
This is a PyBullet version of the HalfCheetah, that is, a task in which a 2-dimensional cheetah is manipulated by continuous control to run faster.
The torque of the six joints can be controlled ($\mathcal{A} = [-1,1]^6$), and the semantic space is a 26-dimensional space.
We collected 64$\times$64 images captured from three different viewpoints for two consecutive timesteps and defined $\mathcal{S}^\tau$ as an image space containing a total of 6 frames.
The total timesteps per episode is fixed to 1000.
The behavior policy to collect $\mathcal{T}^\tau$ is a random policy, and $\mathcal{T}^\tau$ consists of 100 episodes, which is 100k timesteps in total.

In our experiments, information such as $xyz$-coordinates and velocity can be recovered from a combination of multiple images by capturing images from multiple viewpoints at consecutive times, and such a setup is necessary in practice.

\subsection{Setting}
We used a 7-layer convolutional neural network and a 4-layer fully connected neural network for the VAE encoders $E^\tau$ and $\phi$, respectively, for both the proposed and existing methods.
We trained them in gradients using Adam \cite{adam}.
The dimensions of the latent space of VAE $\mathcal{Z}$ were set to 32, 96, and 192 for Shooting, KUKA, and HalfCheetah, respectively.
For CRAR \cite{crar}, we uniformly selected indices $\mathcal{I}$ from $\{i \mid 0\leq i < \lvert \mathcal{T}^\tau \rvert, i\in \mathbb{Z}\}$.
For our method, without an AL setting, $\mathcal{I}$ was selected uniformly and randomly from $\mathcal{O}$.
For Shooting and KUKA, we used a handcrafted policy instead of one trained by RL as $\pi^\sigma$.
In HalfCheetah, we trained $\pi^\sigma$ using PPO \cite{ppo}.

\subsection{Results} \label{sec:exp1}

\begin{table}[t]
\centering
\caption{Results of Shooting. MD values were scaled to $10^2$ for convenience. $\pi^\sigma$ has PP=$45.99$, and the behavior policy has PP=$16.39$.}\label{tab:res1}
\begin{tabular}{l|c|c}
\hline
Method & MD & PP \\
\hline
\multicolumn{3}{c}{$\lvert\mathcal{I}\rvert=0$} \\
\hline
Zhang et al. & $37.42\pm6.23$ & $22.46\pm4.99$ \\
\hline
\multicolumn{3}{c}{$\lvert\mathcal{I}\rvert=10$} \\
\hline
CRAR & $11.00\pm1.72$ & $35.16\pm4.31$ \\
Ours w/o AL & $3.44\pm0.89$ & $43.99\pm1.29$ \\
Ours & $0.16\pm0.12$ & $44.73\pm1.07$ \\
\hline
\multicolumn{3}{c}{$\lvert\mathcal{I}\rvert=50$} \\
\hline
CRAR & $2.80\pm0.70$ & $42.29\pm2.12$ \\
Ours w/o AL & $0.06\pm0.02$ & $46.02\pm0.31$ \\
Ours & $0.02\pm0.00$ & $45.66\pm0.34$ \\
\hline
\end{tabular}
\end{table}

\begin{figure}[t]
\centering
\includegraphics[bb=0 0 720 216, width=0.48\textwidth]{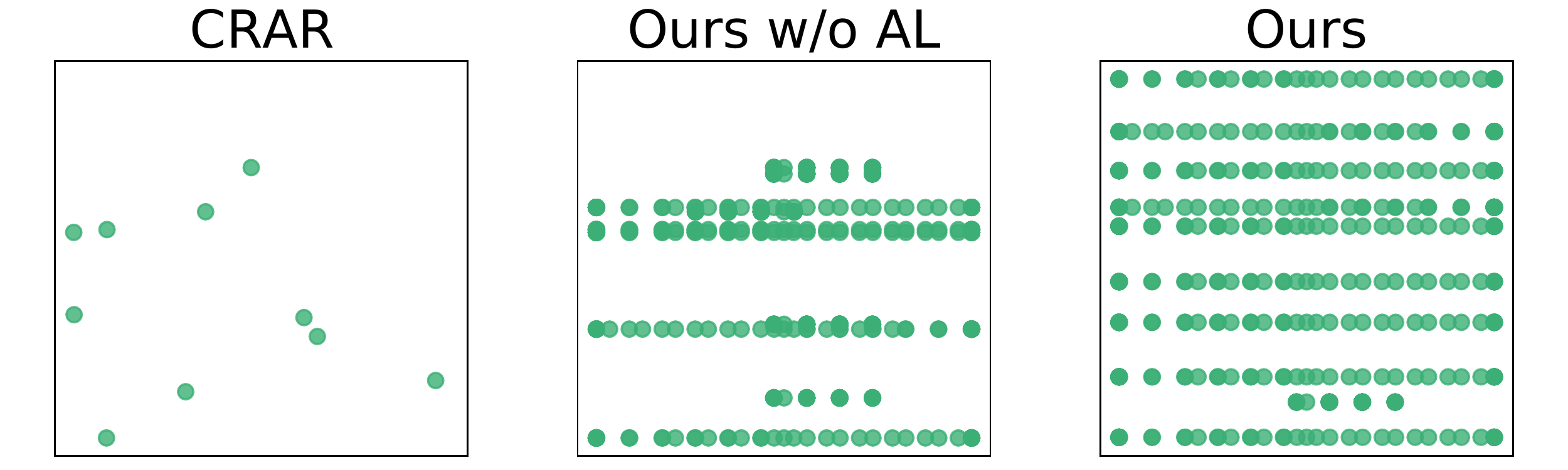}
\caption{Scatter of the obtained semantics on ViZDoom Shooting with $\lvert\mathcal{P}\rvert = \lvert\mathcal{I}\rvert = 10$: $\{F(s^\tau)\mid (s^\sigma,s^\tau)\in\mathcal{P}\}$ for CRAR, and $\{F(s^\tau)\mid (s^\sigma,s^\tau)\in\mathcal{P}\cup\mathcal{P}'\}$ for our method.
Each square represents a 2-dimensional semantic space.
The semantic space shows that both pair augmentation and AL contribute to expanding the coverage.}
\label{fig:illu2}
\end{figure}

\begin{table}[t]
\centering
\caption{Results of KUKA. PP corresponds to grasp success probability. $\pi^\sigma$ has PP=$1.0$, and the behavior policy has PP=$0.048$.}\label{tab:res2}
\begin{tabular}{l|c|c}
\hline
Method & MD & PP \\
\hline
\multicolumn{3}{c}{$\lvert\mathcal{I}\rvert=0$} \\
\hline
Zhang et al. & $0.90\pm0.12$ & $0.12\pm0.08$ \\
\hline
\multicolumn{3}{c}{$\lvert\mathcal{I}\rvert=10$} \\
\hline
CRAR & $0.59\pm0.07$ & $0.24\pm0.22$ \\
Ours w/o AL & $0.35\pm0.05$ & $0.52\pm0.19$ \\
Ours & $0.37\pm0.03$ & $0.65\pm0.07$ \\
\hline
\multicolumn{3}{c}{$\lvert\mathcal{I}\rvert=100$} \\
\hline
CRAR & $0.32\pm0.02$ & $0.52\pm0.15$ \\
Ours w/o AL & $0.11\pm0.01$ & $0.76\pm0.09$ \\
Ours & $0.12\pm0.01$ & $0.90\pm0.04$ \\
\hline
\end{tabular}
\end{table}

\begin{table}[t]
\centering
\caption{Results of HalfCheetah. $\pi^\sigma$ has PP=$2735.73$, and the behavior policy has PP=$-1230.01$.}\label{tab:res3}
\begin{tabular}{l|c|c}
\hline
Method & MD & PP \\
\hline
\multicolumn{3}{c}{$\lvert\mathcal{I}\rvert=0$} \\
\hline
Zhang et al. & $3.71\pm0.32$ & $-1511.97\pm192.51$ \\
\hline
\multicolumn{3}{c}{$\lvert\mathcal{I}\rvert=10$} \\
\hline
CRAR & $1.80\pm0.09$ & $-1411.83\pm144.21$ \\
Ours w/o AL & $0.40\pm0.04$ & $596.20\pm121.43$ \\
Ours & $0.37\pm0.05$ & $580.21\pm71.95$ \\
\hline
\multicolumn{3}{c}{$\lvert\mathcal{I}\rvert=50$} \\
\hline
CRAR & $0.88\pm0.05$ & $-818.29\pm300.08$ \\
Ours w/o AL & $0.12\pm0.01$ & $878.79\pm88.52$ \\
Ours & $0.07\pm0.01$ & $968.49\pm99.53$ \\
\hline
\end{tabular}
\end{table}

\Cref{tab:res1,tab:res2,tab:res3} show the results of the image-to-semantics learning in the three environments.
These tables show the results on average$\pm$std over five trials.
$\lvert\mathcal{I}\rvert$ denotes the number of paired data, which is the annotation cost.
Because of the transition and reward conditions, the PP of $\pi^\sigma\circ F$ on $\mathcal{M}^\tau$ assimilate to that of $\pi^\sigma$ on $\mathcal{M}^\sigma$.

Note that most image-to-image methods shown in \Cref{tab:related} cannot be compared with image-to-semantics methods because some assumptions cannot be satisfied under image-to-semantics settings.
One way to speculate on the performance of the image-to-image techniques in an image-to-semantics setting is to see Zhang et al. \cite{zhang}.
Zhang et al. used domain adaptation \cite{adda}, which is commonly used in image-to-image learning; thus, their method can be interpreted as a representative example in which the techniques cultivated in image-to-image are imported to image-to-semantics.
Although CycleGAN \cite{CycleGAN2017} is also widely employed in image-to-image learning, along with domain adaptation, they confirmed in their experiments that this method did not outperform their method in the image-to-semantics setting \cite{zhang}.

In all cases, compared with the approach of Zhang et al. \cite{zhang}, our approaches with and without AL achieved a smaller MD and a greater PP.
Zhang et al.’s approach is designed to learn $\hat{F}$ without a paired dataset to eliminate the annotation cost.
However, learning without pairs does not necessarily lead to the true image-to-semantics translation mapping, as observed in the high MD and low PP in our results.
This result shows the effectiveness of the paradigm using paired data when aiming for higher performance policy transfer while compromising the annotation cost to prepare a small number of paired data.

By comparing the results of our approaches with and without AL and those of CRAR, we confirmed the efficacy of pair augmentation in achieving a smaller MD and higher PP. 
We achieved PP=$44.73\pm1.07$ in Shooting with 10 pairs using pair augmentation and AL, which is more than the PP=$42.29\pm2.12$ achieved by CRAR with 50 pairs.
In addition, in KUKA, we achieved PP=$0.65\pm0.07$ in 10 pairs, which exceeds PP=$0.52\pm0.15$ achieved by CRAR with 100 pairs. This means that the annotation cost was reduced by more than $\times$5 and $\times$10, respectively.
This difference is even more pronounced in HalfCheetah. This may be because the ratio $\lvert \mathcal{P} \cup \mathcal{P}'\rvert / \lvert \mathcal{P}\rvert$ is the greatest in this environment: CRAR uses $\lvert \mathcal{P}\rvert = \lvert \mathcal{I}\rvert$ paired data, whereas our proposed approach used an additional $\lvert \mathcal{P}' \rvert = 999 \lvert \mathcal{I}\rvert$ augmented paired data because the number of timesteps per episode was 1000 in this environment.

A tendency of reduced MD and increased PP was observed in the proposed approach with AL compared to that without AL.
Specifically, AL reduced MD except in KUKA, and clearly improved PP in KUKA, while achieving competitive PP in the other two environments.

In \Cref{fig:illu2}, we present the effectiveness of our approach in Shooting.
The proposed AL maximized the diversity in the latent space that represents the image space, but the diversity was also maximized when this result was visualized in the semantic space.

\subsection{Experiments with Errors}
In this section, we verify the robustness of the proposed method against errors in annotation and state transitions.

\subsubsection{Annotation Error}
In our previous discussion and in experiments of \Cref{sec:exp1}, we assumed that we could query oracle $F$, that is, true image-to-semantics mapping by human annotations.
However, because human annotation indicates the process of assigning semantics to images by humans, errors are expected to occur in the output semantics.
Therefore, we provided a new experimental setup here: for some $s^\tau \in \mathcal{S}^\tau$, we can observe $F(s^\tau) + \epsilon$ instead of $F(s^\tau)$ while creating the paired dataset $\mathcal{P}$, where $\epsilon \in \mathbb{R}^{\dim(\mathcal{S}^\sigma)}$ is a random vector representing the annotation error.

\subsubsection{Transition Error}
In reality, the state transition function on the simulator $Tr^\sigma$ is expected to contain modeling errors.
For example, environment parameters such as friction coefficients and motor torques in the real world cannot be accurately estimated in the simulator, and thus, state transitions in reality cannot be correctly imitated.
Therefore, we provided a new experimental setup here: for some $(s,a) \in \mathcal{S}^\sigma\times\mathcal{A}$, we could obtain $Tr^\sigma(s,a) + \epsilon$ instead of $Tr^\sigma(s, a)$ while augmenting a paired dataset, where $\epsilon \in \mathbb{R}^{\dim(\mathcal{S}^\sigma)}$ is a random vector representing the transition error.
Note that when training $\pi^\sigma$, we used the one without errors in our experiments.

\subsubsection{Error Generation}
We generated two types of errors by adding a random variable $\epsilon \in \mathbb{R}^{\dim(\mathcal{S}^\sigma)}$.
Here, we denoted a value of the $h$-th dimension of $x \in \mathbb{R}^H$ as $x_{(h)} \in \mathbb{R}$.
We sampled $\epsilon_{(h)} \sim \mathcal{N}_{(h)}$, where $\mathcal{N}_{(h)}$ is a Gaussian distribution with mean $\mu=0$ and standard deviation $\sigma=\alpha \cdot {\rm std}[s^\sigma_{(h)}]_{s^\sigma \in \mathcal{T}^\sigma}$.
Here, ${\rm std}[s^\sigma_{(h)}]_{s^\sigma \in \mathcal{T}^\sigma}$ is the sample standard deviation of a source trajectory $\mathcal{T}^\sigma$ collected by a behavior policy in source MDP, and $\alpha \geq 0$ is the noise scale.

For the annotation error, using the semantics sequence of augmented pairs $\{\widehat{s}_{i+t}^\sigma\}_{t \in \mathcal{E}_{i}}$ provided without error, we provided the semantics as $F(s^\tau_i) + \bar{\epsilon}_i$ for $\mathcal{P}$ and $\{\widehat{s}_{i+t}^\sigma + \bar{\epsilon}_i\}_{t \in \mathcal{E}_{i}}$ for $\mathcal{P}'$, where $\bar{\epsilon}_i$ is a realized random vector with $\alpha > 0$.
For the transition error, we defined $\{\widehat{s}_{i+t}^\sigma + \sum_{j=1}^{t} \bar{\epsilon}_{i,j}\}_{t \in \mathcal{E}_{i}}$ for $\mathcal{P}'$, where $\bar{\epsilon}_{i,j}$ is the realized random vector.
Note that, here, we approximated the error generation based on the following assumption: $Tr^\sigma(s,a) = s + Tr^\sigma_{\Delta}(a)$, that is, $Tr^\sigma(s+\bar{\epsilon}_1,a)+\bar{\epsilon}_2 = s+Tr^\sigma_{\Delta}(a)+\bar{\epsilon}_1+\bar{\epsilon}_2$.
This is an approximation simplifying implementation; however, for Shooting and KUKA, the above assumption is actually satisfied for almost all states and actions.

\subsubsection{Results}

\begin{table}[t]
\centering
\caption{Results with annotation errors.}\label{tab:noisy-anno}
\begin{tabular}{l|c|c|c}
  \hline
Method & $\alpha$ & MD & PP \\

\hline
\multicolumn{4}{c}{Shooting ($\lvert\mathcal{I}\rvert = 50$)} \\
\hline

 CRAR & \multirow{2}{*}{0.0} & $2.80\pm0.70$ & $42.29\pm2.12$ \\
 Ours w/o AL &  & $0.06\pm0.02$ & $46.02\pm0.31$ \\
\hline
CRAR & \multirow{2}{*}{0.04} & $2.99\pm0.93$ & $42.91\pm1.71$ \\
Ours w/o AL &  & $0.12\pm0.05$ & $45.90\pm0.54$ \\
\hline
CRAR & \multirow{2}{*}{0.15} & $3.50\pm1.21$ & $43.51\pm2.04$ \\
Ours w/o AL &  & $0.47\pm0.07$ & $44.56\pm0.59$ \\
\hline
CRAR & \multirow{2}{*}{0.3} & $4.90\pm0.85$ & $40.83\pm2.78$ \\
Ours w/o AL &  & $1.42\pm0.15$ & $42.47\pm1.74$ \\

\hline
\multicolumn{4}{c}{KUKA ($\lvert\mathcal{I}\rvert = 100$)} \\
\hline

CRAR & \multirow{2}{*}{0.0} & $0.32\pm0.02$ & $0.52\pm0.15$ \\
Ours w/o AL &  & $0.11\pm0.01$ & $0.76\pm0.09$ \\
\hline
CRAR & \multirow{2}{*}{0.04} & $0.33\pm0.03$ & $0.58\pm0.11$ \\
Ours w/o AL &  & $0.12\pm0.01$ & $0.76\pm0.18$ \\
\hline
CRAR & \multirow{2}{*}{0.15} & $0.32\pm0.03$ & $0.53\pm0.15$ \\
Ours w/o AL &  & $0.14\pm0.01$ & $0.68\pm0.15$ \\

\hline
\multicolumn{4}{c}{HalfCheetah ($\lvert\mathcal{I}\rvert = 50$)} \\
\hline

CRAR & \multirow{2}{*}{0.0} & $0.88\pm0.05$ & $-818.29\pm300.08$ \\
Ours w/o AL &  & $0.12\pm0.01$ & $878.79\pm88.52$ \\
\hline
CRAR & \multirow{2}{*}{0.04} & $0.91\pm0.03$ & $-919.86\pm328.14$ \\
Ours w/o AL &  & $0.12\pm0.01$ & $787.0\pm230.32$ \\
\hline
CRAR & \multirow{2}{*}{0.15} & $0.99\pm0.06$ & $-833.35\pm464.76$ \\
Ours w/o AL &  & $0.15\pm0.02$ & $616.63\pm260.73$ \\
\hline

\end{tabular}
\end{table}

Here, we analyze the effect of two types of errors on $\mathcal{P}$ and $\mathcal{P}'$, and understand how this affects the approximation of $F$.
Therefore, we do not experiment with the method of Zhang et al., which does not utilize paired datasets.
In addition, to eliminate the effect of the choice of $\mathcal{I}$ on the generation of $\mathcal{P}$ and $\mathcal{P}'$ in the comparison between the proposed method and CRAR, we conducted experiments under the setting without AL.

The results with annotation errors are shown in \Cref{tab:noisy-anno}.
In both CRAR and the proposed method, the semantics of the paired data deviated from the true data as the scale of the annotation error $\alpha$ increased; thus, we observed that MD tends to increase for both methods.
Although PP tended to decrease only for the proposed method, the proposed method achieved better MD and PP than CRAR for the same error scale $\alpha$.
In addition, we confirmed that PP with an annotation error for our method remains comparable to the case without an annotation error for a certain degree of $\alpha$.
For example, in KUKA experiments, the proposed method achieved PP $= 0.76\pm0.18$ with $\alpha=0.04$, which was close to PP $= 0.76\pm 0.09$ without annotation error.
We conclude that the proposed pair augmentation is effective in image-to-semantics learning even in the presence of annotation errors.

\begin{table}[t]
\centering
\caption{Results with transition errors.}\label{tab:noisy-tran}
\begin{tabular}{l|c|c|c}
  \hline
Method & $\alpha$ & MD & PP \\

\hline
\multicolumn{4}{c}{Shooting ($\lvert\mathcal{I}\rvert = 50$)} \\
\hline

  CRAR & \multirow{2}{*}{0.0} & $2.80\pm0.70$ & $42.29\pm2.12$ \\
  Ours w/o AL &  & $0.06\pm0.02$ & $46.02\pm0.31$ \\
\hline
Ours w/o AL & 0.01 & $0.12\pm0.04$ & $45.90\pm0.51$ \\
\hline
Ours w/o AL & 0.04 & $0.67\pm0.12$ & $44.78\pm1.16$ \\
\hline
Ours w/o AL & 0.1 & $3.43\pm1.00$ & $43.44\pm1.60$ \\

\hline
\multicolumn{4}{c}{KUKA ($\lvert\mathcal{I}\rvert = 100$)} \\
\hline

CRAR & \multirow{2}{*}{0.0} & $0.32\pm0.02$ & $0.52\pm0.15$ \\
Ours w/o AL &  & $0.11\pm0.01$ & $0.76\pm0.09$ \\
\hline
Ours w/o AL & 0.01 & $0.12\pm0.01$ & $0.80\pm0.11$ \\
\hline
Ours w/o AL & 0.04 & $0.14\pm0.00$ & $0.67\pm0.22$ \\

\hline
\multicolumn{4}{c}{HalfCheetah ($\lvert\mathcal{I}\rvert = 50$)} \\
\hline

CRAR & \multirow{2}{*}{0.0} & $0.88\pm0.05$ & $-818.29\pm300.08$ \\
Ours w/o AL &  & $0.12\pm0.01$ & $878.79\pm88.52$ \\
\hline
Ours w/o AL & 0.01 & $0.18\pm0.02$ & $426.61\pm221.12$ \\
\hline
Ours w/o AL & 0.04 & $1.21\pm0.13$ & $-523.8\pm336.51$ \\
\hline

\end{tabular}
\end{table}

The results with transition errors are shown in \Cref{tab:noisy-tran}.
Note that CRAR does not use $Tr^\sigma$; thus, the result did not depend on transition error scale $\alpha$; the result of CRAR with $\alpha > 0$ matched the result of $\alpha=0$.
Because the proposed pair augmentation scheme used $Tr^\sigma$ to generate semantics, for larger $t \in \mathcal{E}_i$, the variance of error was expected to be large; then, the augmented semantics in $\mathcal{P}'$ were far from the actual semantics.
In fact, we observed an increase in MD and a decrease in PP in the proposed method as the scale of $\alpha$ increased.
In contrast, both MD and PP were better than CRAR up to $\alpha=0.04$ for Shooting and KUKA, and up to $\alpha=0.01$ for HalfCheetah.
This indicates that the proposed pair augmentation is effective in reducing the annotation costs up to a certain level of transition errors.

\subsection{Effect of Behavior Policy}
To further reveal the behavior of image-to-semantics methods, we evaluated them on HalfCheetah by adopting a low performance policy, rather than the random policy, as the behavior policy.
We pre-trained the low performance policy with a small number of iterations using PPO.

We observed that, compared with \Cref{tab:res3} and \Cref{tab:res4}, the performance of the behavior policy affected the PP of the resulting target agent.
Here, the PP of the random policy was $-1230.01$ and that of the low performance policy was $822.39$.
Therefore, the PP of our proposed method was improved from $968.49\pm99.53$ to $1527.37\pm133.19$ when $\lvert\mathcal{I}\rvert=50$.

\begin{table}[t]
\centering
\caption{Results of HalfCheetah when $\mathcal{T}^\tau$ was collected by the low performance policy. The behavior policy has PP=$822.39$ and $\pi^\sigma$ has PP=$2735.73$. The trajectory for calculating MD is collected by the low performance policy.}\label{tab:res4}

\begin{tabular}{l|c|c}
\hline
Method & MD & PP \\
\hline
\multicolumn{3}{c}{$\lvert\mathcal{I}\rvert=0$} \\
\hline
Zhang et al. & $1.03\pm0.19$ & $-1241.31\pm523.06$ \\
\hline
\multicolumn{3}{c}{$\lvert\mathcal{I}\rvert=10$} \\
\hline
CRAR & $0.74\pm0.06$ & $-1549.82\pm106.04$ \\
Ours w/o AL & $0.09\pm0.02$ & $1117.05\pm127.93$ \\
Ours & $0.06\pm0.00$ & $1145.77\pm110.29$ \\
\hline
\multicolumn{3}{c}{$\lvert\mathcal{I}\rvert=50$} \\
\hline
CRAR & $0.37\pm0.04$ & $-1416.37\pm194.31$ \\
Ours w/o AL & $0.03\pm0.00$ & $1393.11\pm226.39$ \\
Ours & $0.02\pm0.00$ & $1527.37\pm133.19$ \\
\hline
\end{tabular}
\end{table}

These results indicate that owing to the low performance of the random policy, faster-running states, that is, states with high velocity, cannot be observed; in other words, the random policy can only observe a limited state. 
This limitation could lead to an increase in the approximation error of $\hat{F}$.
This implied that image-to-semantics is affected by the performance of the behavior policy in some tasks.

A promising result for the image-to-semantics framework is that the target agents obtained by our approach outperform the behavior policy.
In particular, in \Cref{tab:res4}, the PP of the behavior policy is $822.39$; furthermore, when image-to-semantics was performed with 50 annotations ($\lvert\mathcal{I}\rvert=50$), we obtained a PP of $1527.37\pm133.19$.
In other words, we could achieve a higher performance compared with that of the behavior policy using a small number of annotations and the image-to-semantics protocol.

In the previous discussion, we found that the PP achieved by the image-to-semantics framework is affected by the quantity and quality of paired data, and the region of state space comprising the dataset for training $\hat{F}$.
In fact, as an extreme example, $\hat{F}$ trained using $\lvert\mathcal{P}\rvert$=100k with the trajectories collected by the optimal policy, achieved PP=$2624.65\pm29.09$, which is almost identical to PP=$2735.73$, the performance of the optimal source policy.
Note that such a near-complete policy transfer is already achieved in Shooting and KUKA, as shown in \Cref{tab:res1,tab:res2}.

\section{Conclusion}

In this study, we investigated the image-to-semantics problem for vision-based agents in robotics.
Using paired data for learning image-to-semantics mapping is favorable for achieving high-performance policy transfer; however, the cost of creating paired data cannot be ignored. 
This study contributes to existing literature by reducing the annotation cost using two techniques: pair augmentation and active learning. We also confirmed the effectiveness of the proposed method in our experiments.

In future work, we must address the following limitations: (1) Experiments have not been conducted using actual robots; therefore, it is not known how difficulties specific to actual robots will affect the image-to-semantics performance; (2) We cannot always freely query $Tr^\sigma$; therefore, it would be beneficial to know if we can substitute the one learned using source trajectory, similar to \cite{zhang,worldmodels}; (3) In some cases, the transition error is too large, and we would like to be able to improve the approximation accuracy of $\hat{F}$ by considering performing pair augmentation for $\{t \mid t \in \mathcal{E}_i, t \leq K\}$ rather than $\mathcal{E}_i$. This expectation is because augmented semantics with larger $t\in \mathcal{E}_i$ are inaccurate.
Furthermore, we would like to find a way to automatically determine such a $K$.

\bibliographystyle{IEEEtran}
\bibliography{rl}

\end{document}